\documentclass[a4paper,11pt]{article}

\usepackage{physics, amssymb}
\usepackage{color, bm, soul}
 \usepackage{graphicx}
\usepackage{url}
\usepackage[affil-it]{authblk}
\usepackage{amsmath}
\usepackage{wrapfig}
\usepackage{algorithm, algpseudocode}
\usepackage[T1]{fontenc}
\usepackage{times}
\usepackage{float}
\usepackage{hyperref}
\usepackage{fancyhdr}

\begin{document}

\title{A Stochastic Birth-and-Death Approach for Street Furniture Geolocation in Urban Environments}

\author{Evan Murphy, Marco Viola, Vladimir A. Krylov}
\affil{School of Mathematical Sciences, Dublin City University}
\date{}
\maketitle
\thispagestyle{fancy}

\begin{abstract}
In this paper we address the problem of precise geolocation of street furniture in complex urban environments, which is a critical task for effective monitoring and maintenance of public infrastructure by local authorities and private stakeholders. To this end, we propose a probabilistic framework based on energy maps that encode the spatial likelihood of object locations. Representing the energy in a map-based geopositioned format allows the optimisation process to seamlessly integrate external geospatial information, such as GIS layers, road maps, or placement constraints, which improves contextual awareness and localisation accuracy. A stochastic birth-and-death optimisation algorithm is introduced to infer the most probable configuration of assets. We evaluate our approach using a realistic simulation informed by a geolocated dataset of street lighting infrastructure in Dublin city centre, demonstrating its potential for scalable and accurate urban asset mapping. The implementation of the algorithm will be made available in the GitHub repository \href{https://github.com/EMurphy0108/SBD_Street_Furniture}{https://github.com/EMurphy0108/SBD\_Street\_Furniture}.

\end{abstract}

\section{Introduction and Related Works}
\subsection{Problem Description and Overview}
Precise detection and geolocation of street furniture, such as road signs, telegraph poles, street lights, is an important tool for the rapid development of modern urban environments. Local councils, smart cities, and telecommunication companies are just some of the organisations that require accurate and up-to-date information for asset managing purposes. Many deep-learning pipelines have been proposed for the detection and geolocation of object from street level and satellite imagery, either independently, or as part of a multi-view framework~\cite{sun, lefevre}. Whilst these methods have displayed good performance, the process of accurately locating positive detections is inevitably subject to noise. The aim of this work is to develop a methodology with the view to improve the accuracy of the geolocation.

The focus of this work will be the construction of an energy map, incorporating both detection and geographic information system (GIS) based information. Such an energy map should be designed in such a way that favourable object locations are reflected through low energy values, and unfavourable locations through high energy values. A stochastic birth \& death optimisation scheme will be discussed to identify the object locations which minimise the overall energy. The optimisation scheme that is discussed follows a simulated annealing type schedule, governed by the Gibbs distribution defined by the constructed energy.

\subsection{Related Work}

In the past, probable object locations have been modelled as pairwise intersections of camera-to-object rays resulting from positive individual detections~\cite{krylov}. This space of intersections, together with unary and pairwise energy terms, is modelled as a Markov Random Field (MRF), and forms the basis for a quadratic pseudo-boolean optimisation problem. This approach has been shown to be effective at reducing duplicate detections and false positives, and achieving high accuracy and object recall rates. Performance faces limitations, however, due to lack of flexibility in proposed object locations, and absence of support for GIS-based information. Efforts have been made to adapt this formulation to incorporate such information~\cite{mrfosm}, but drawbacks with respect to flexibility still remain. These, in particular, pertain to the formulation of the MRF optimization which is applied to the irregular grid of intersections, rather than explores geolocations in the more native 2D map format.

Convolutional neural networks (CNN) have also proven effective for the tasks of object detection and geolocation. In particular, in conjunction with Markovian point processes~\cite{mpp}, these types of models allow for the incorporation of image data and spatial interactions between objects. Specifically, geographical information arising from satellite imagery can be leveraged to produce well-informed object location proposals. Object locations and interactions are derived from an energy based on information obtained from a CNN, through its corresponding Gibbs density. This CNN approach does however, require extensive training (dependant on availability of large-scale reliable data) to deliver the best performance.

A schematic for simulating such a point process has also been explored in the form of a stochastic birth \& death (SBD) schematic~\cite{sbd}. This formulation employs a simulated annealing type cooling schedule to an energy-based model to extract the configuration of objects of minimal energy. The construction of the energy is such that probable object locations are encoded through low energy locations. Convergence of the SBD process to the minimiser of the energy is proven, and a time-discretisation of the algorithm is proposed.

\section{Stochastic Geometrical Model}

\subsection{Birth \& Death Formulation}

Consider a bounded domain $V \subset \mathbb{R}^2$. Define $\Gamma_n(V) \subset\mathbb{R}^2 \times \mathbb{R}^+$ to be the collection of all $n$-point configurations of objects. In this context, an object is a disc defined by a centre-point in $\mathbb{R}^2$, and some radius mark in $\mathbb{R}^+$. 
In this paper, the radius mark will be allowed to vary across objects, giving some information on positional uncertainty. Set $\Gamma(V) = \bigcup_{n=1}^\infty \Gamma_n(V)$ to be the collection of all configurations with finitely many points. This work proposes to build an energy function $H: \Gamma(V) \to \mathbb{R}$, which will be used to guide a birth \& death optimisation schematic, similar to that proposed in~\cite{sbd}. The energy of each point $(\bm{x},r)\in g$, where $g\in \Gamma(V)$, will be comprised of a unary term $U(\bm{x},r)$, and a pairwise interaction term $I(\bm{x},r; g)$, with the energy of the entire configuration $g$ being the sum of the energies of the individual configuration points. The unary energy is derived from a static energy map, the construction of which is discussed in the next subsection. The pairwise interaction energy of a configuration point is proportional to its area of overlap with other configuration points.

The optimisation scheme used will be based on a simulated annealing type algorithm with temperature cooling, and governed by the Gibbs distribution. That is, the probability of the state of the system being the configuration $g \in \Gamma(V)$ is given as
\begin{align*}
    P_\text{Gibbs}(g; \beta) = \frac{1}{Z}e^{\beta H(g)}.
\end{align*}
Here, $\beta$ is the inverse temperature parameter, and $Z$ is an (often intractable) normalising constant. This work appeals to the results shown in~\cite{sbd}, namely that as $\beta \to \infty$, the distributions $P_\text{Gibbs}(g;\beta)$ converge weakly to the distribution $\sum_{g^* \in S} c_{g^*}\delta_{g^*}$, where $S \subset \Gamma(V)$ is the set of global minimisers of $H$ of minimal size, $\delta_{g^*}$ denotes a Dirac distribution centred at a global minimiser $g^* \in S$, and constants $\{ c_{g^*}:g^* \in S\}$ are non-negative and sum to $1$. The birth \& death formulation is proposed as a less computational intensive alternative to the traditional Reversible Jump Markov Chain Monte Carlo, for sampling from distributions of varying dimension. \\

We now describe a time-discretised implementation of the stochastic birth \& death process, which is controlled by the corresponding simulated annealing cooling schedule. The entire process is shown in Algorithm~\ref{alg:birth_death_sa}. 

Over the area of interest, define a grid of pixels of size $h \times w$. Each pixel is identified by its coordinate taking values in $\mathcal{D}=\{1,\cdots,h\}\times\{1,\cdots,w\}$. Configurations of points of the form $(\bm{x}, r)$ are considered, where $\bm{x}\in \mathcal{D}$, and $r \in \mathbb{Z}_{>0}$. Denote by $\Gamma(\mathcal{D})$ the collection of all such configurations of finitely many points. 
Given an unary energy map $D\in\mathbb{R}^{h \times w}$, the algorithm generates a sequence of configurations $\{g_m : m=1,2,\ldots\} \subset \Gamma(\mathcal{D})$. The generation of the next configuration $g_{m+1}$ from the current configuration $g_m$ is described below. It suffices to specify the parameters $N_0 >0, \varepsilon \in (0,1),$ and $\beta >1$. Note that the parameter $\varepsilon$ models the time-discretisation of the continuous birth \& death process. $\beta$ is the inverse temperature parameter, and controls the cooling schedule of the simulated annealing.\\

\textbf{Birth Step.} At the $m^\text{th}$ birth step, we sample the number of points $N_m$ from a Poisson distribution ($N_m \sim \text{Pois}(N_0)$), and the radius $r_m$ from an exponential distribution ($r_m \sim [\text{Exp}(\frac{1}{10})]_{2,\cdots,10}$). Note that the chosen radius is rounded to the nearest integer, and resampled until it falls within the appropriate range of values. $N_m$ point coordinates $(\bm{x}_1, \cdots, \bm{x}_{N_m})$ are sampled with radius $r_m$ from $\mathcal{D}$ and added to the configuration, according to the probability distribution given by
\begin{align*}
    P(\bm{x};r) = \frac{U(\bm{x},r)}{\sum\limits_{\bm{y}\in \mathcal{D}} U(\bm{y},r)}, \quad \forall \bm{x} \in \mathcal{D}.
\end{align*}

\textbf{Death Step.} Sort the current configuration points $\hat{g}_{m+1} = g_m \cup \{\bm{x}_1, \cdots, \bm{x}_{N_m} \}$ with respect to energy, highest to lowest. For each point $(\bm{x},r)$ in this order, compute the probability of death as
\begin{align*}
    d(\bm{x},r) = \frac{\varepsilon^m a(\bm{x,r})}{1+\varepsilon^m a(\bm{x},r)}, \text{ where } a(\bm{x},r) = \exp(\beta^{m}[H(\hat{g}_{m+1})-H(\hat{g}_{m+1}\backslash\{(\bm{x},r)\})]).
\end{align*}
The point $(\bm{x},r)$ is removed from the configuration with probability $d(\bm{x},r)$ before computing the probability of death for the next point. The surviving points are returned in the form of the next configuration, $g_{m+1}$.\\

\textbf{Convergence.} At each iteration, the total energy of the configuration is calculated, and compared against the minimal energy seen so far. If the algorithm, has not improved over a fixed number of iterations, the algorithm is considered to have converged, and the configuration of minimal energy is returned.\\

\begin{algorithm}[t]
\caption{Stochastic Birth \& Death Algorithm}
\label{alg:birth_death_sa}
\begin{algorithmic}[1]

\State Set domain $\mathcal{D}$, parameters $N_0, \varepsilon, \beta$, and $T_\text{wait}$. Set initial configuration $g_0 = \emptyset$, minimal energy $H_\text{min} = 0$, and counters $T=0$, $m=0$. Generate unary energy map $D$.
\While{$T<T_\text{wait}$}
    \Statex \textsc{Birth Step:}
    \State Sample $N_m \sim \text{Poisson}(N_0)$, Sample $r_m \in \{2, \dots, 10\}$ with $P(r_m = k) \propto \text{Exp}\left(-\frac{1}{10}k\right)$ 
    \State Sample $N_m$ points $\{x_i:i = 1,\cdots,N_m \} \subset \mathcal{D}$ with $P(x_i; r_m) = U(x_i, r_m)/\sum_{y \in D} U(y, r_m)$
    \State $\hat{g}_{m+1} \gets g_m \cup \{(x_1, r_m), \dots, (x_{N_m}, r_m)\}$
    \Statex \textsc{Death Step:}
    \State Sort $\hat{g}_{m+1}$ by energy (descending: from worst to best) 
    \For{each point $(x, r)$ in sorted $\hat{g}_{m+1}$ (in order)}
        \State $a(x, r) \gets \exp(\beta^{m}[H(\hat{g}_{m+1})-H(\hat{g}_{m+1}\backslash\{(x,r)\})])$
        \State $d(x, r) \gets \varepsilon \cdot a(x, r) / (1 + \varepsilon \cdot a(x, r))$
        \State Remove $(x, r)$ from $\hat{g}_{m+1}$ with probability $d(x, r)$
    \EndFor
    \State $g_{m+1} \gets \hat{g}_{m+1}$ 
    \Statex \textsc{Convergence Check:}
    \If{$H(g_{m+1}) < H_{min}$}
        \State $H_\text{min} \gets H(g_{m+1})$, $g^* \gets g_{m+1}$, $T \gets 0$
    \Else
        \State $T \gets T + 1$
    \EndIf
    \State $m \gets m + 1$
\EndWhile
\State \Return $g^*$
\end{algorithmic}
\end{algorithm}

\subsection{Construction of Energy Function}
The availability of street-camera locations, together with positive individual detections is assumed. To each detection, an estimate of distance, bearing, and measure of confidence is given. Camera-to-object rays are projected according to the compass bearings, and pairwise intersections of rays occurring within 20 metres from their respective cameras are marked. The camera-to-intersection distance along the ray is also computed.

To each intersection of camera-to-object rays, confidence measures $c_1,c_2 \in (0.5,1)$, depth estimates $d_1,d_2 \in \mathbb{R}^+$, and camera-to-intersection distances $\Delta_1, \Delta_2 \in [0,20]$ can be assigned, arising from the two original individual detections respectively. Confidence measures typically arise from classification methods (CNNs) as a measure of certainty (softmax class weight) about any specific detection. Depth estimates can be obtained either though monocular depth estimation from individual street level images, or from cameras with active sensors aboard, such as LiDAR. Each intersection can  thus be identified by the tuple $I = (i,j,c_1,c_2, d_1, d_2, \Delta_1, \Delta_2)$, where $(i,j) \in \mathcal{D}$ is the two-dimensional location coordinate of the intersection.

\begin{figure}[H]
    \begin{center}
    \includegraphics[scale=0.3]{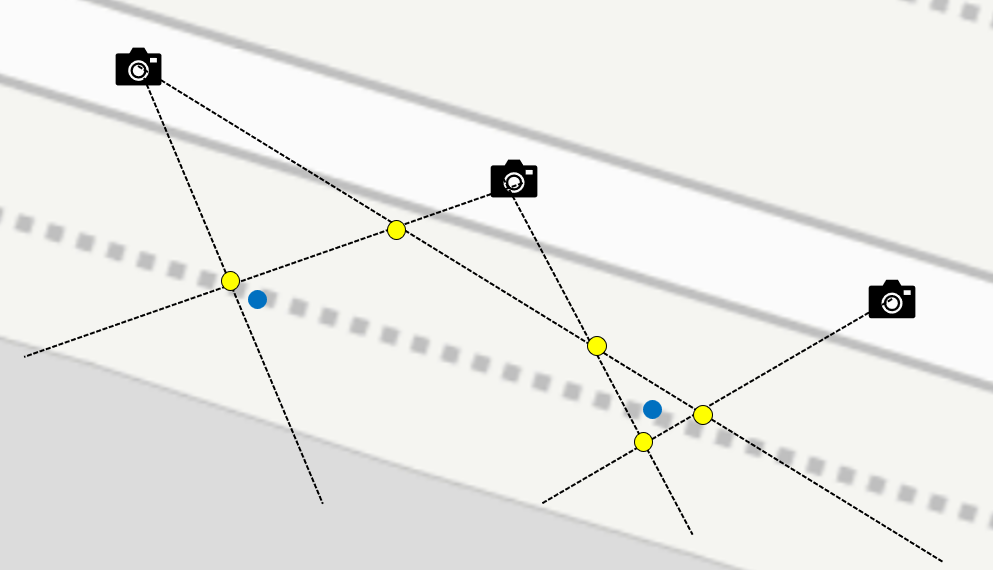}\quad \includegraphics[scale=0.282]{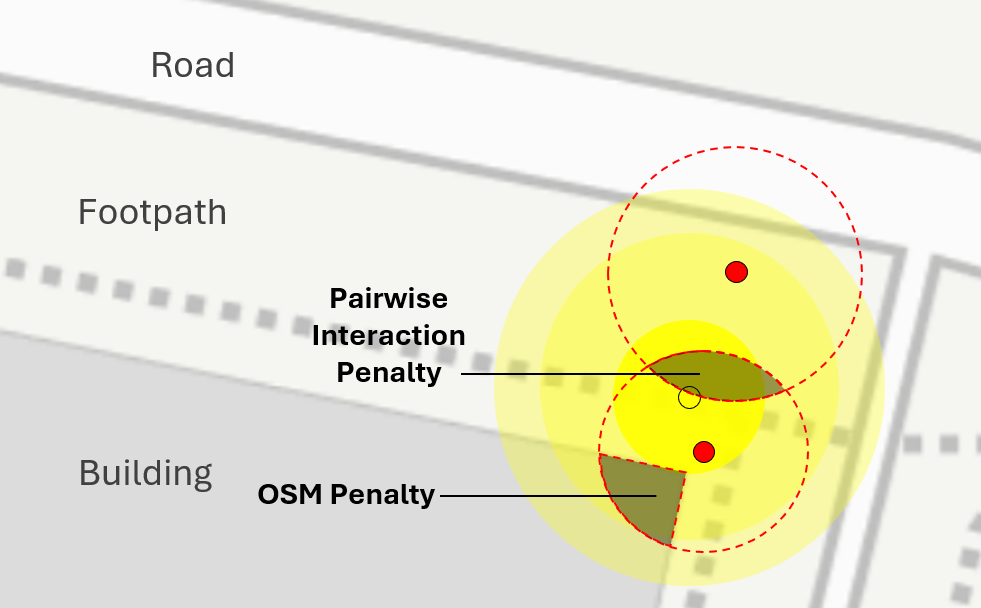}
    \caption{Left: Examples of intersections (yellow) of object-to-camera rays arising from detections of the real objects (blue). Right: Two points (red) proposed by the SBD algorithm. Confidence and depth consistency energy is dispersed from an intersection (yellow). The lower object receives a GIS-penalty for overlap with a building. Both objects receive a pairwise interaction penalty.}
    \label{fig:4}
    \end{center}
\end{figure}

Let $\mathcal{I} = \{ I^{(k)}\}_{k=1}^N$ be the collection of all $N$ available pairwise intersections, where
\begin{align*}
I^{(k)} = \left(i^{(k)},j^{(k)},c_1^{(k)},c_2^{(k)}, d_1^{(k)}, d_2^{(k)}, \Delta_1^{(k)}, \Delta_2^{(k)}\right).
\end{align*}
For each $k=1,\cdots,N$, define $s_1^{(k)} = c_1^{(k)} c_2^{(k)}$ and $s_2^{(k)} = \norm{d_1^{(k)}-\Delta_1^{(k)}}+\norm{d_2^{(k)}-\Delta_2^{(k)}}$ to be the confidence and depth consistency scores of intersection $k$ respectively.
Moreover, let $G^{(k)}$ denote a Gaussian kernel with $\sigma^{(k)} \propto \frac{1}{d_1^{(k)}} + \frac{1}{d_2^{(k)}}$. This formulation of standard deviation reflects the inherent increase in uncertainty arising from close-range detections. Define $M^{(k)} \in \mathbb{R}^{h \times w}$ by
\begin{align*}
    M^{(k)}_{x,y} = \left\lbrace\begin{array}{ll}
        1, & \text{ if } (x,y) = (i^{(k)}, j^{(k)}),\\
        0, & \text{ otherwise.}
    \end{array}\right.
\end{align*}
Then a weighted map of confidence and depth consistency energies can be constructed as 
\begin{align*}
    \sum\limits_{k=1}^N \left(w_1 s_1^{(k)} + w_2 s_2^{(k)} \right)G^{(k)} \ast M^{(k)},
\end{align*}
where $\ast$ denotes the standard kernel-matrix convolution operation. A GIS map $R \in \mathbb{R}^{h\times w}$ is also defined:
\begin{align*}
R_{x,y} = \left\lbrace\begin{array}{ll} 1, &\text{ if the pixel is occupied by infrastructure},\\
0, &\text{otherwise}.
\end{array}\right.
\end{align*}
This term can inform the geolocation process of the occupancy of any given location by incompatible objects, like buildings, water bodies, etc. In this paper we employ Open Street Maps (OSM) to provide such information. The final unary energy map $D \in \mathbb{R}^{h \times w}$ is given as
\begin{align*}
D = \sum\limits_{k=1}^N (w_1 s_1^{(k)} + w_2 s_2^{(k)} )G^{(k)} \ast M^{(k)} + w_3 R .
\end{align*}
The energy function $H:\Gamma(\mathcal{D}) \to \mathbb{R}$ will consist of a unary component, and a pairwise interaction component. Given two configuration points $(\bm{x}_1, r_1), (\bm{x_2}, r_2)$, let $A[(\bm{x}_1, r_1), (\bm{x_2}, r_2)]$ denote the area of overlap of the discs with centres $\bm{x}_1, \bm{x}_2$, and radii $r_1, r_2$, respectively. The energy function $H$ can now be defined as
\begin{align*}
    H(g) = \sum\limits_{(\bm{x},r)\in g} \left[ U(\bm{x},r)  + \alpha \sum\limits_{(\bm{x'},r')\in g\backslash\{(\bm{x},r)\}} \frac{A[(\bm{x},r),(\bm{x'},r')]}{\pi r'^2}\right], \text{where } U(\bm{x},r) = 
    \sum\limits_{{\bm{y}\in\mathcal{D},\;\norm{\bm{x}-\bm{y}}\leq r}} D_{\bm{y}}
\end{align*}
for any $g \in \Gamma(\mathcal{D})$, where $U(\bm{x},r)$ is the unary energy of a point $(\bm{x},r)$ and $\alpha$ is a weighting between the unary and pairwise interaction energy terms. It suffices to choose the weighting parameters $w_1,w_2,w_3 \in \mathbb{R}$. A conceptual visualization of the camera-to-object rays, and computation of unary and pairwise energy terms is presented in Figure~\ref{fig:4}.

\section{Experimental Validation}
\begin{figure}[!t]
    \begin{center}
    \includegraphics[scale=0.35, trim={0 4.5cm 0 0},clip]{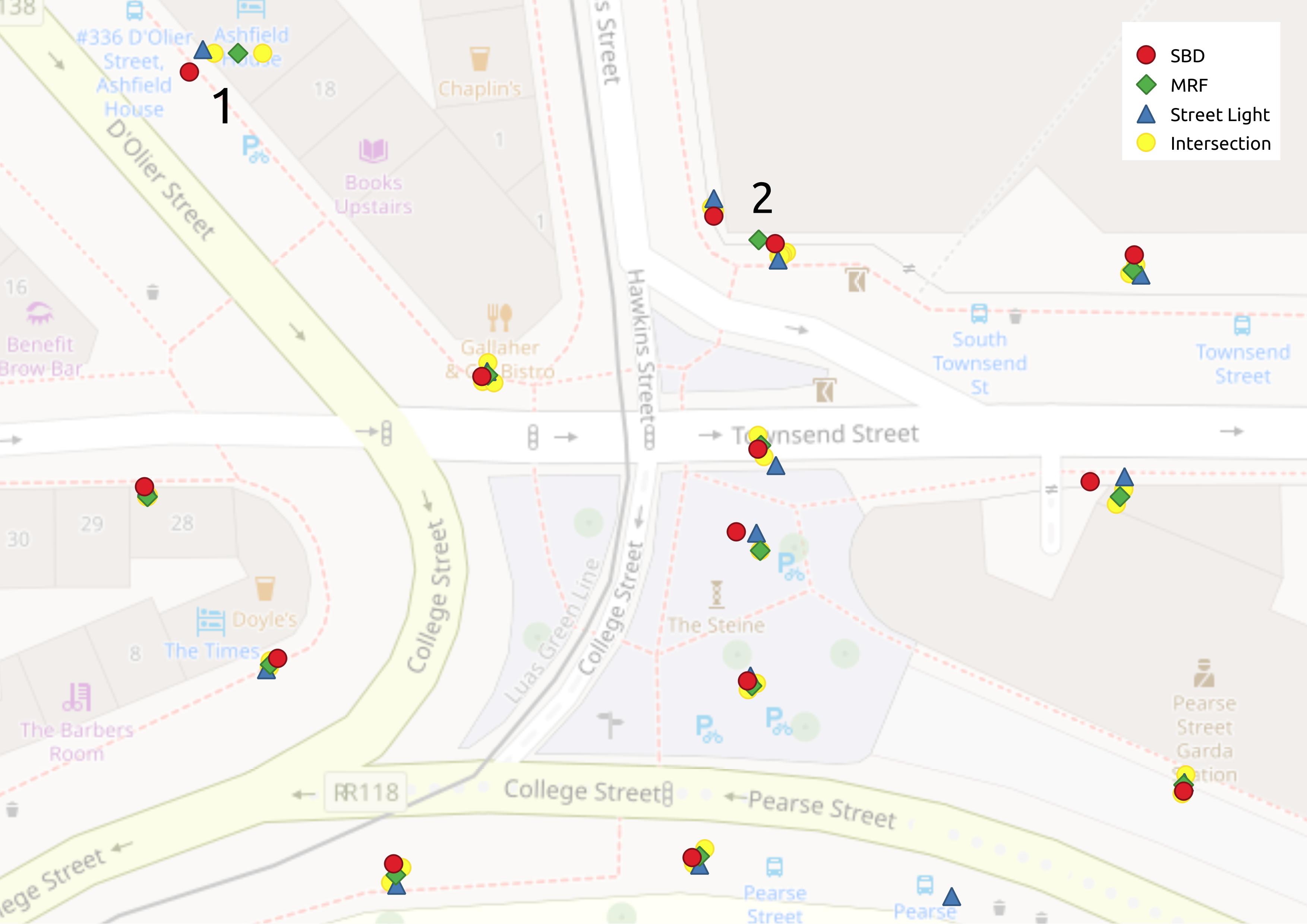}
    \caption{Example of detections of street lights (blue) by SBD (red) and MRF (green), guided by pairwise intersections of camera-to-object rays (yellow). SBD detections are represented by object centre. These samples come from a simulation with noise level $1$.}
    \label{fig:5}
    \end{center}
\end{figure}

This section focuses on the quantitative performance of the SBD implementation. We consider both the quality of the results, as well as their stability. Due to the stochastic nature of the algorithm, it is important to assess the reliability and reproducibility of results. Experimentation was conducted in an area of central Dublin, with a total of $680$ ground-truth objects (street lights) to be detected. A synthetic dataset of individual street-level detections was constructed. An example of detections made by the SBD algorithm is shown in Figure~\ref{fig:5}. Each simulation of the SBD algorithm was run on a standard desktop computer, for approximately $4000$ iterations and an average runtime of around $4$ minutes.

\subsection{Simulation of Data}
This work assumes the availability of individual detections of street furniture. For the purposes of experimentation, and comparison with state-of-the-art, such data has been simulated. Although the use of simulated data has limitations, we believe that it is appropriate for evaluating our model in the first instance. Having direct control over the noisiness of the data allows us to determine on which real-world data our model would be appropriate to use. We also hope to mitigate the issue to limited availability of real-world data. Geographic coordinates of street camera locations were obtained from the official Google Street View API in central Dublin. The geographic coordinates of street lights (objects) were obtained from~\cite{dcc}. These datasets have been used to simulate individual detections of objects from street cameras, with added noise and contaminations.

For each camera location, the $15$ nearest objects were identified, and camera-to-object distances computed. Given an object of distance $d$ from the camera, a positive detection was declared with probability $0.7$, if $d \in [0\text{m}, 2\text{m})\cup (10\text{m},20\text{m})$, and probability $0.9$ if $d \in [2\text{m},10\text{m}]$. Otherwise, the object was not marked as detected. These probabilities were chosen to reflect the inherent uncertainty in real-world detections. For each positive detection with camera-to-object distance $d$, and compass bearing $B$, noise was added by sampling from the Normal distribution: $d_\text{noise} \sim \text{N}(d, \sigma_1)$ and $B_\text{noise} \sim \text{N}(B,\sigma_2)$. The location of the detection was calculated to be a distance $d_\text{noise}$ from the camera location, with bearing $B_\text{noise}$. 

Contaminate detections were also simulated such that the dataset mimic the real detections typically obtained from a CNN with a fraction of false positives. Given $N$ detections, and a contamination level of $p$, $n = \lfloor pN \rfloor$ contamination points were generated. To this effect, $n$ camera locations were chosen uniformly and without replacement. A camera-to-object distance $d$, and bearing $B$, were sampled from the uniform distributions $\text{Unif}(1\text{m},15\text{m})$ and $\text{Unif}(0^\circ, 360^\circ)$ respectively. The locations of these contaminate detections were calculated at a distance $d$ from the chosen camera location, with bearing $B$. Each contaminating detection was then treated as an object, and detections from nearby cameras were simulated. To every simulated detection, a CNN measure of confidence was assigned, of the form $1-\alpha$, where $\alpha$ is sampled from an exponential distribution. This choice imitates empirical observations that can be made from real-world data.
For the purposes of simulation, four levels of noise were considered, see Table \ref{tab:1}. Note that noise level $1$ serves a baseline realistic case, whilst levels $0$ and $3$ are the cases of optimistic (low) and pessimistic (significant) noise respectively.

\begin{table}
    \begin{center}
    \begin{tabular}{|c|c|c|c|c|c|c|c|c|c|}
         \hline
         & \multicolumn{3}{c|}{Parameters} & \multicolumn{2}{c|}{Within Cluster Distances} & \multicolumn{2}{c|}{Object Count} & \multicolumn{2}{c|}{Distance to GT}\\
         \hline
         Noise Level & $\sigma_1$ & $\sigma_2$ & $p$ & Median & SD & Median & SD & Median & SD\\
         \hline
         0 & $1$m & $2^\circ$ & $0.03$ & $1.718$m & $0.930$m & $654$ & $9.23$ & $1.374$m & $0.919$m\\
         1 & $2$m & $3^\circ$ & $0.05$& $1.103$m & $0.910$m & $593$ & $8.08$ & $1.316$m & $0.896$m\\
         2 & $3$m & $4.5^\circ$ & $0.075$ & $1.030$m & $0.870$m & $547$ & $4.93$ & $1.568$m & $1.020$m\\
         3 & $4$m & $6^\circ$ & $0.1$ & $1.000$m & $0.944$m & $502$ & $4.903$ & $1.833$m & $1.145$m\\
         \hline
    \end{tabular}
    \caption{Description of the four noise levels considered and details of the stability of the SBD algorithm at each ($10$ simulations). Rightmost columns report details of average detection-to-ground truth (GT) distance.}
    \label{tab:1}
    \end{center}
\end{table}

\subsection{Stability}
\begin{figure}[H]

\begin{center}
    \includegraphics[scale=0.55]{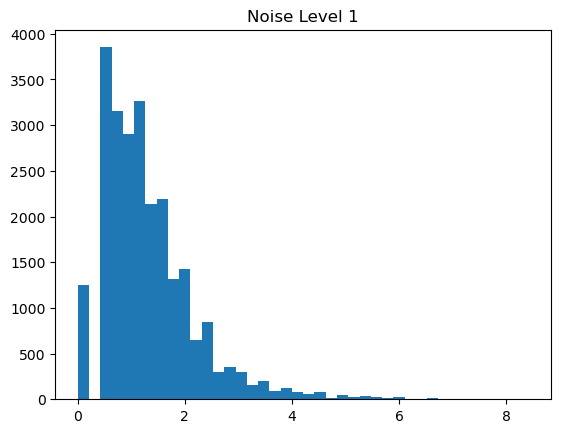} \includegraphics[scale=0.55]{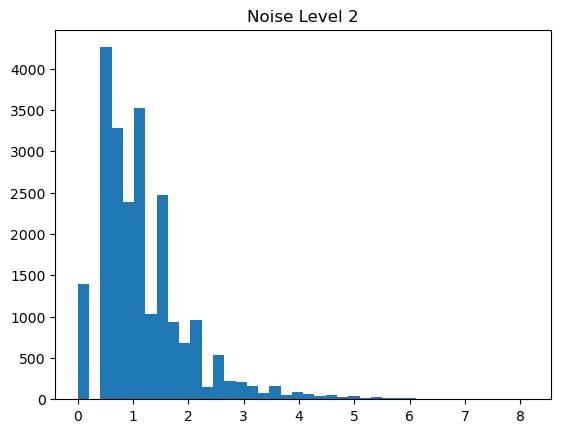}
\end{center}
\vspace{-.5cm}
\caption{Histograms of within cluster distances for noise levels $1$ and $2$.}
\label{fig:1}
\end{figure}

Due to the stochastic nature of the employed optimisation, the exact position of the objects changes every time the SBD procedure is run. We now examine the stability of the birth \& death algorithm, by running $10$ simulations under identical settings, for each of the considered four noise levels. Both positional and object count stability is considered. Positional stability is measured by considering clusters of generated points corresponding to a single ground-truth object. Within each cluster, pairwise distances between points are calculated. The median and standard deviation (SD) of distance for each noise level is summarised in Table \ref{tab:1}. There is a noticeable decrease in distance as noise present in the data increases. For low noise settings, street-level detections are clustered tightly, and SBD experiences more pairwise interactions, which introduces volatility in the final result. For higher noise settings, these detections are more sparse. This lends itself to a simpler optimisation problem, meaning SBD produces more consistent (although not necessarily more accurate) results. Histograms for the within cluster pairwise distances for noise levels $1$ and $2$ are displayed in Figure \ref{fig:1}. These histograms are heavily right-skewed, indicating low within cluster distances, and a stable output from SBD.\\

Object-count stability is also evaluated by considering the number of objects in the optimal configuration produced as a result of the $10$ simulations at each noise level. The median and standard deviation of the number of objects is reported in Table \ref{tab:1}. These results once again imply that SBD is more stable for high noise levels.

\subsection{Performance}
\begin{figure}[!t]
    \centering
    \includegraphics[scale=0.525]{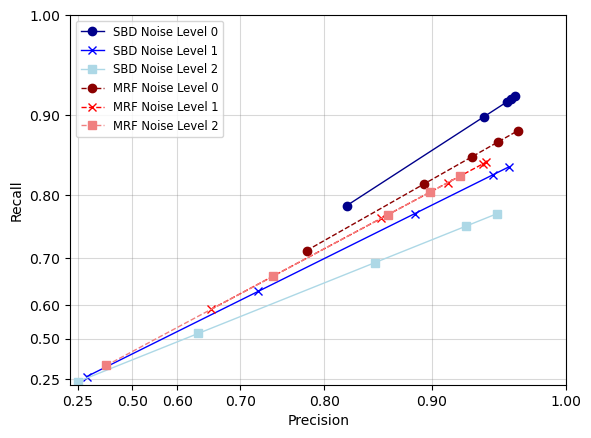} \qquad \includegraphics[scale=0.525]{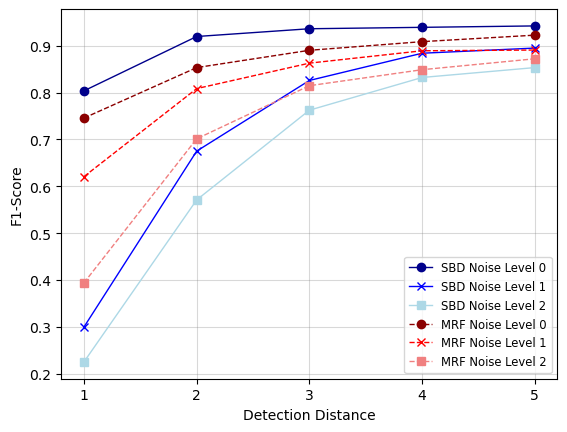}
    \vspace{-.25cm}
    \caption{Plot of precision v. recall (left) and F1-score vs. detection distance (right) for a single simulation of SBD and MRF at different noise levels. The left plot employs non-linear scaled axes for better visualisation.}
    \label{fig:2}
\end{figure}

The numerical performance of this model is measured through precision and recall against the dataset of ground-truth object locations. Weighting parameters used are $w_1 = -3, w_2 = 0.1, w_3 = 0.4$ and $\alpha = 10$, selected through manual empirical tuning. Sufficient exploration of configurations was encouraged by choosing $N_0 = 100$ and $T_\text{wait} = 500$. The simulated annealing parameters were set to $\varepsilon = \beta = 0.999$ to ensure an appropriate cooling schedule.  As a baseline, this performance is compared against an off-the-shelf implementation of the \textit{MRF with clustering} model described in \cite{krylov}. Visualisations of these results are given in Figure \ref{fig:2}. Precision and recall are computed for both SBD and MRF results, at distances from $1$ to $5$ metres from ground-truth object locations, for three distinct noise levels (left). From these results, one can observe that SBD outperforms MRF for the lowest level of noise. Moreover, observations of detection to ground truth distance given in Table \ref{tab:1} show that the model performs well, especially given a detection tolerance of least $2$ metres. A plot of F1-score against detection distance for SBD and MRF at each noise level is also provided (right). Again, it can be observed that for a low level of noise, SBD shows good performance, greater than that of MRF. Note that due to the randomness present in the simulation of detections, some ground-truth objects remain undetectable by both models, effectively capping recall.

Visually, the advantages of SBD over MRF can be seen in Figure \ref{fig:5}. Cluster $\bm{1}$ shows the benefit of the additional GIS information. The SBD point is located outside of the building, closer to the ground-truth location. In contrast, MRF located based solely on the two intersections, resulting in a much less accurate result. Cluster $\bm{2}$ exhibits a case where SBD discerned two ground-truth objects, whilst MRF proposed a single detection.

\section{Conclusions}

We introduced a probabilistic geolocation method for urban street furniture based on a geopositioned energy map and a stochastic birth-and-death (SBD) optimisation algorithm. The experiments demonstrate that the proposed SBD formulation can match, and in some cases outperform, the established (non-stochastically optimized) MRF-based model in terms of localisation accuracy. In addition to its competitive performance, the method benefits from a flexible map-based representation that allows seamless integration of external geospatial information such as GIS data, road networks, and placement constraints. These features make it particularly attractive for real-world applications in infrastructure monitoring and management. 
Future work will explore scaling to larger urban areas, incorporating multimodal data sources, and further developing the stochastic optimisation technique, with the goal of improving convergence speed, robustness to noise, and scalability.

\section*{Acknowledgements}
This research was supported by the ADAPT Research Centre, which is funded by Science Foundation Ireland (Grant 13/ RC/2106\_P2) and is co-funded by the European Regional Development Fund.

{
\small
\bibliographystyle{apalike}
\bibliography{imvip}
}

\end{document}